\documentclass[10pt,conference,twocolumn]{ieeetrans}
\usepackage[version=4]{mhchem}
\usepackage{siunitx}
\usepackage{longtable,tabularx}
\usepackage[utf8]{inputenc}
\usepackage{amsmath}
\usepackage{graphicx}
\usepackage{subcaption}
\usepackage{mathtools, cuted}
\usepackage{setspace} 
\usepackage{hyperref}
\usepackage{import}
\usepackage{pdfpages}
\usepackage{amsfonts}

\IEEEoverridecommandlockouts 

\graphicspath{ {./images/} }
\title{Motion Planning for Safe Landing of a Human-Piloted Parafoil}
\author{Maximillian Fainkich, Kiril Solovey, and Anna Clarke
\thanks{M.F.\ and A.C.\ are with the Stephen B. Klein Faculty of Aerospace Engineering, and K.S.\ is with the Viterbi Faculty of Electrical and Computer Engineering, 
        Technion--Israel Institute of Technology, 3200003 Haifa, Israel. 
        {\tt\small maxf1000@gmail.com, kirilsol@technion.ac.il, anna.clarke@technion.ac.il}}
}

\begin{document}

\maketitle
\thispagestyle{plain}
\pagestyle{plain}

\begin{abstract}
    Most skydiving accidents occur during the parafoil-piloting and landing stages and result from human lapses in judgment while piloting the parafoil. Training of novice pilots is protracted due to the lack of functional and easily accessible training simulators.  Moreover, work on parafoil trajectory planning suitable for aiding human training remains limited. To bridge this gap, we study the problem of computing safe trajectories for human-piloted parafoil flight and examine how such trajectories fare against human-generated solutions. For the algorithmic part, we adapt the sampling-based motion planner Stable Sparse RRT (SST) by Li et al., to cope with the problem constraints while minimizing the bank angle (control effort) as a proxy for safety. We then compare the computer-generated solutions with data from human-generated parafoil flight, where the algorithm offers a relative cost improvement of 20\%-80\% over the performance of the human pilot. We observe that human pilots tend to, first, close the horizontal distance to the landing area, and then address the vertical gap by spiraling down to the suitable altitude for starting a landing maneuver. The algorithm considered here makes smoother and more gradual descents, arriving at the landing area at the precise altitude necessary for the final approach while maintaining safety constraints. Overall, the study demonstrates the potential of computer-generated guidelines, rather than traditional rules of thumb, which can be integrated into future simulators to train pilots for safer and more cost-effective flights.
\end{abstract}

\section{Introduction}
\label{intro}

Despite the increasing reliability of modern parachute systems, skydiving accidents still occur, mainly due to human errors and lack of piloting skills.  In fact, \(60\%\) of parachute fatalities in 2023 in the United States were due to piloting errors on perfectly functioning parachutes \cite{uspa}. These errors result from a lack of experience among novice jumpers and from a variety of factors that influence parachute dynamics (e.g., temperature, humidity, elevation, wind, shape and material of parachutes, and terrain), thereby prolonging the acquisition of piloting skills. Unlike aircraft pilots, parafoil pilots learn to operate the parafoil in real-time in the air (starting from their very first flight) due to the lack of adequate simulators, and are trained according to rules of thumb. As part of the process to develop simulators that help parafoil pilots acquire the essential experience, an adequate trajectory leading to the target landing area must be computed and displayed to trainees. Computer-generated approaches may also provide better solutions for landing trajectories than the aforementioned rules of thumb, as well as better principles for pilot training. In addition, future augmented-reality devices may also allow real-time feedback and instruction provided mid-flight, actively helping with collision avoidance.

\begin{figure}[t]
\centering
\includegraphics[width=0.7\linewidth]{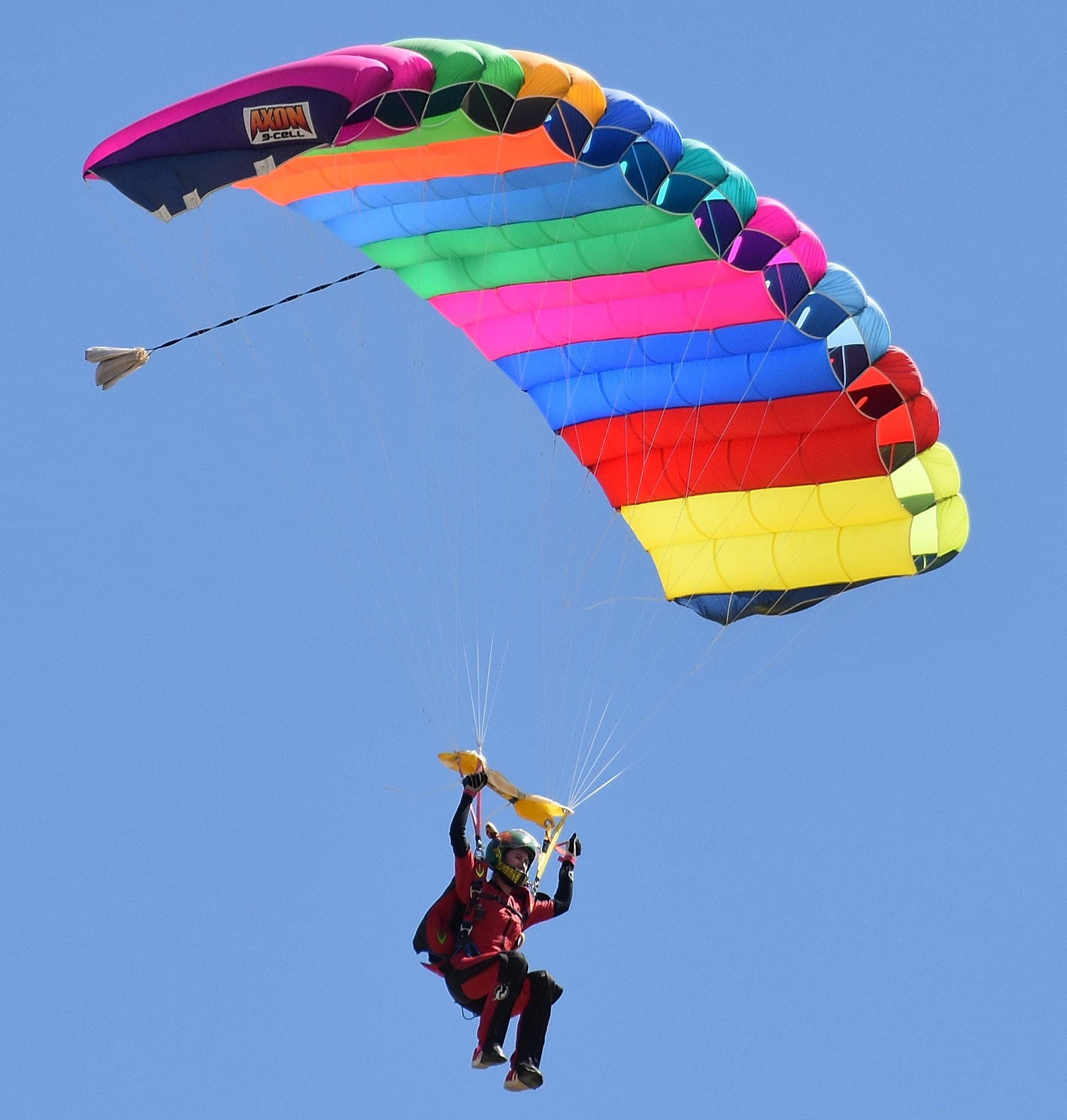}
\caption{A parafoil pilot during data gathering for this work.}
\label{annapicture}
\end{figure}

Many solutions to the problem of an autonomous parafoil have been proposed using a control approach. A full control stack is designed in \cite{fullstackpara}, with the guidance being split into three phases. The first phase is the homing phase, where the parafoil brings itself into the general vicinity of the target. The second phase is the energy management phase, where the parafoil burns off altitude to prepare itself for landing. The final stage is the flare maneuver, where the parafoil heads to the target point against the wind. Worthy of note is this algorithm's similarity to human pilot guidance strategies, which will be covered later on in this work. Another work \cite{simpletermialguidance} solves the terminal guidance problem by calculating the trajectory to bring the parafoil through a carefully timed maneuver, with the turn times being calculated to counteract the impact of the wind. Model Predictive Control (MPC) is then used to follow the trajectory, as well as a simplified control law with a yaw rate tracking control. In \cite{optimalparafoil}, the problem is framed as a two-point boundary value problem (BVP) and solved using optimal control. Solutions are generated for minimum-time  and minimum-control-effort optimization. An algorithm based on the excess altitude left for maneuvering decides in real-time which solution to use. Optimal control can also be employed~\cite{highwindoptimalcontrol} with a focus on cases with high winds. The wind is handled by moving to a coordinate frame aligned with its direction, which simplifies the dynamics of the problem. While control approaches provide simple-to-implement solutions, they struggle with abstract constraints such as obstacles and do not offer global optimality guarantees, especially in the presence of nonholonomic dynamic constraints of the parafoil.

Other approaches, different from the control solution, have dealt with constraints more successfully. In ~\cite{parafoilbeziercurve} Bezier curves are used in order to solve the BVP. The solution trajectories are computed under constraints of obstacle collision avoidance and yaw rate limits, and are optimized using a cost of the altitude at the target location ("altitude error"). If one Bezier curve is not sufficient for finding a solution to the problem (the altitude at the target is not low enough) then two can also be used and stitched together, and if no solution can be found then a circular loitering path or Dubins Path is followed. However, there are no theoretical guarantees as to the existence of a solution. Another work relies on dynamic programming \cite{parafoildynamicprogramming}, where the trajectory is built by a separation into equally spaced sections based on path length and optimization over each section, albeit with a very simple kinematic model and a large computation time. The cost function used for the optimization is the distance from the target point. Reinforcement learning (RL) can be used \cite{rlparafoil} to generate control inputs to a parafoil of unknown dynamics. The algorithm is fed images of its surroundings and must perform obstacle recognition and avoidance after being trained with a reward function based on keeping hazards out of the center of the images. Another RL-based approach~\cite{rlparafoil2} uses  a reward function based on the miss distance and the angle against the wind in the final approach. The returned solution is then smoothed out along the trajectory to make implementation easier. Terminal guidance is solved in \cite{mcrobustguidance} by performing a Monte Carlo analysis on a number of different trajectories and then choosing the trajectory most robust to uncertainties such as wind. While such approaches can better deal with constraints, they offer no guarantees on feasibility or solution quality. 

A small number of works explore the use of sampling-based methods for parafoil motion planning, which account for state and control constraints and rely on general parafoil dynamic models. In \cite{optRRTPara}, a simple RRT algorithm with some sampling optimizations is used to solve the problem for a hilly environment. Rewire-RRT~\cite{Rewire} is implemented to solve the problem with a cost function based on the predicted miss distance from the target, combined with some safety function for preventing collisions. In order to calculate controls for propagation and new node generation, the control necessary to connect two nodes in 2D space is calculated and then applied in the 3D space. In the case of rewiring, the distance from the new node to the original node is checked to validate that it is under a certain boundary. In ~\cite{dynamicPathPlanning}, RRT is used at each step to generate the next section of the trajectory. The wind and parafoil dynamics are updated at each step, and then used to generate the tree with a cost function based on distance to the target and control smoothness. In \cite{CC-RRT1,CC-RRT2} the CC-RRT algorithm is used to propagate uncertainty distributions to account for unknowns in the problem such as the wind. An accurate wind model is developed and the algorithm is tested against the model to show improved performance under wind uncertainty. In addition, a cost-to-go function is implemented to select the best path before planning completes, enabling execution and planning to run in parallel. While solving the problem for an unmanned parafoil, the trajectories computed in these papers are not suitable for human pilots due to a lack of human-specific safety constraints and no direct optimization of the control effort.

The focus of research to date has been on resolving the challenges of autonomous flight, which differ from those faced by human pilots. Autonomous flight of parafoils involves high levels of uncertainty and limited control, making it difficult to generate exact solutions. These problems are less pronounced in human parafoil flight as a human pilot offers better adaptability and can think of creative solutions on the fly. Instead, human parafoil flight must place stricter bounds on the safety of said human pilot, thus forcing more constraints on the problem. For example: landing against the direction of the wind, avoiding flying above objects that produce turbulence (such as buildings), and avoiding flying in proximity to other pilots.

\vspace{5pt} \noindent \textbf{Contribution.} In this work, we study the problem of computing safe trajectories for human-piloted parafoil flight and assessing the quality of human-generated solutions. To achieve that, we first adapt the sampling-based motion planner Stable Sparse RRT (SST)~\cite{SST}, to cope with the problem constraints while minimizing the bank angle (control effort) as a proxy for safety. Minimization of the control effort inherently results in fewer and slower turns, preventing the most dangerous for a human pilot maneuver: a low-altitude sharp turn. Through in-depth case studies, we then compare the computer-generated solutions with real-life data gathered from GPS sensors attached to human skydivers. We observe that our solution matches the general characteristics of human-piloted flight,  while bringing relative cost improvements over a human pilot of $20\%-80\%$. These findings highlight the potential of incorporating computer-generated solutions into the pilots' training process, and serve as a stepping stone toward more advanced approaches that plan for a team of human pilots. Such solutions can be used to train human pilots for safer and lower control effort flight patterns. 

The rest of this work is organized as follows. The mathematical model of the parafoil flight is presented in Sec. \ref{model}, followed by the problem formulation in Sec. \ref{problem}. The solution to the problem is discussed in Sec. \ref{method}, followed by a performance analysis of said solution in Sec. \ref{results}. Comparisons with real flight data are shown in Sec. \ref{realdata}. Finally, concluding remarks are presented in Sec. \ref{conc}.

\section{Parafoil Kinematics}
\label{model}

\begin{figure}[b]
\centering
\includegraphics[width=0.8\linewidth]{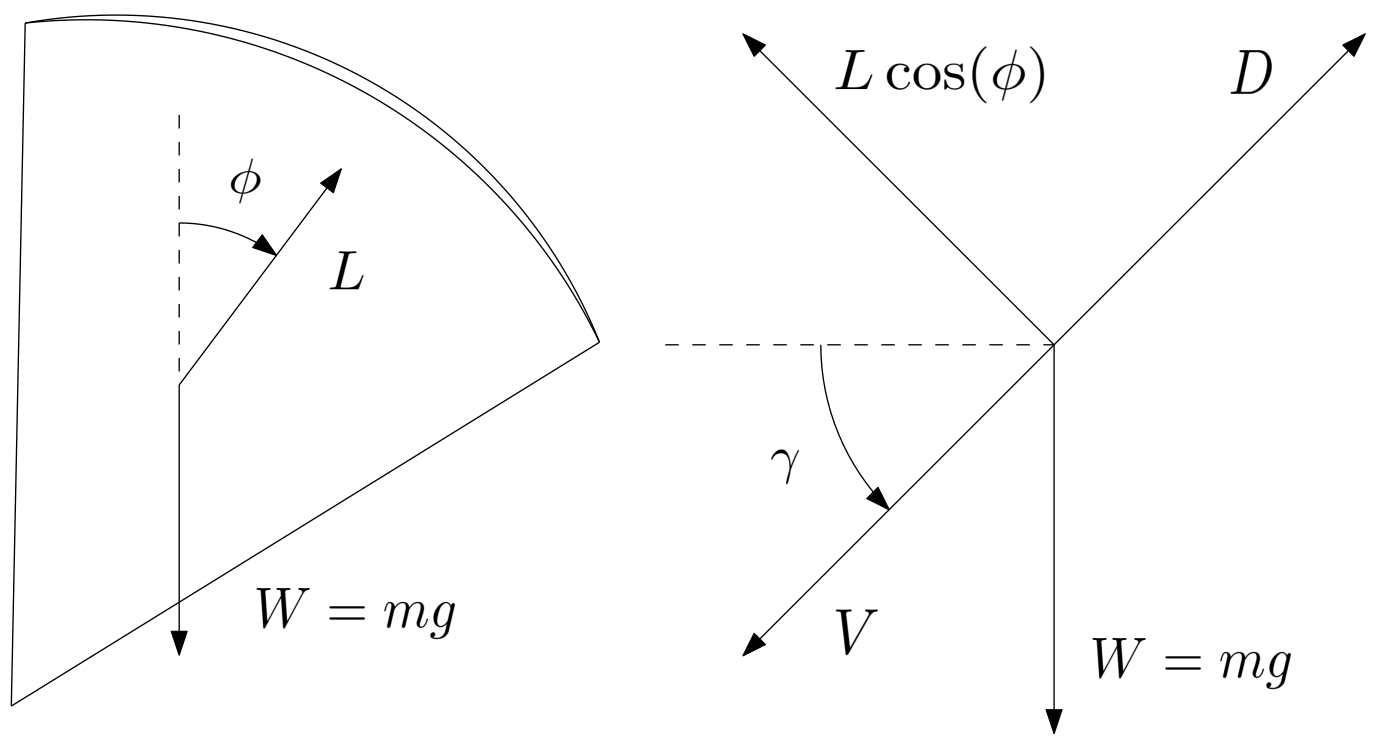}
\caption{Parafoil force diagram: (left)  frontal view and (right) lateral view.} 
\label{kinematics}
\end{figure}

We consider the parafoil kinematics presented in~\cite{pini2}. These kinematics were chosen due to their simplicity and ease of implementation while still being accurate enough to provide feasible results. Schematics of parafoil flight are given in Fig. \ref{kinematics}. This model assumes perfect knowledge of the state and surroundings as well as no time delay in the control value. The state vector of the parafoil is given by
\begin{equation}
\mathbf{x} = [x\;\; y\;\;h\;\;\psi]^T, 
\end{equation}
where \(x\) is the current x-coordinate of the parafoil, \(y\) is the y-coordinate, \(h\) is the current altitude, and \(\psi\) is the parafoil's heading. The equations of motion for the simplified model can then be written as 
\begin{equation}
\label{dynamics}
    \begin{cases}
        \dot{x}&=V\cos\gamma\cos\psi+w_x \\
        \dot{y}&=-V\cos\gamma\sin\psi+w_y \\
        \dot{h}&=V\sin\gamma+w_h\\
        \dot{\psi}&=-\frac{g}{V}\tan{\phi}
    \end{cases}
\end{equation}
where \(V\) is the parafoil's speed, \(\gamma\) is the flight angle, \(g\) is the acceleration due to gravity, \(\phi\) is the bank angle of the parafoil, and \([w_x, w_y, w_h]\) is the vector of the wind. The parafoil will be assumed to be in straight glide or equilibrium turning glide, which implies \(\dot{V}=\dot{\gamma}=\ddot{\psi}=0\). The flight angle depends on the bank angle which redirects the lift vector of the parafoil, changing the balance of forces in the vertical direction. Solving for the flight angle from the force diagram shown in Fig. \ref{kinematics} under the equilibrium assumption gives the expression
\begin{equation}
\gamma = \arctan\left(\frac{-1}{\cos(\phi)\cdot L/D}\right),
\end{equation}
where \(L/D\) is the glide ratio of the parafoil, also assumed to be constant for simplicity. This leads to a dynamics model where the parafoil is characterized by two constant characteristics---its speed and glide ratio---and the bank angle is used as the system's control input. As mentioned earlier, it is assumed that the bank angle command is implemented ideally and without latency.

\section{Problem Statement}
\label{problem}

Consider a parafoil operating in a bounded three-dimensional workspace \(W \subseteq \mathbb{R}^3\) cluttered with a set of obstacles \(\mathbb{O}\), where every obstacle \(o\in\mathbb{O}\) is a closed subset of \(W\). The parafoil starts with initial conditions \([x_0\in\mathbb{R},y_0\in\mathbb{R},h_0\in\mathbb{R}_{>0},\psi_0\in[-\pi, \pi]]\) at time \(t=t_0\) and individual parameters \(V,L/D\). The parafoil is assumed to have a rigid body \(S\subset W\) and moves according to the dynamics presented in Eq. \eqref{dynamics}. The parafoil has a landing area (goal region) \(X^G\subset W\). The landing time is denoted as \(t=t_{f}\) and only has an upper bound due to the dynamics of the problem (not otherwise constrained).

Given the above, we wish to compute an input function of the bank angle \(\boldsymbol{\phi}:[t_0,t_{f}]\rightarrow [-\pi/6,\pi/6] \)  which brings the parafoil into the designated landing area \(X^G\) without colliding with any obstacles (\(\forall t: S(t)\cap\mathbb{O}=\emptyset\)). This input vector should  minimize the total control effort as defined in the cost function:
\begin{equation} \label{cost_function}
J_{t_f} = \int_{t_0}^{t_f}\boldsymbol{\phi}(t)^2\, \mathrm{d}t.
\end{equation}
This cost function minimizes the amount of turns and the time each turn lasts. The latter prevents the parafoil from considerable increases of vertical speed, thus making the heading adjustments smooth and gentle. The former ensures arrival at the landing area at the precise altitude for entering the final landing approach without the need for further adjustments. In addition, the final approach of the parafoil (the last \(100[m]\) above the ground) is performed straight in the direction against the wind vector \(w\). 
Turns at low altitudes are undesirable due to an inherent increase of vertical velocity, resulting in an insufficient altitude to complete a turning maneuver and return to the parafoil's nominal glide angle required for performing an efficient flare (parafoil stall) for landing. 

\section{Methodology}
\label{method}

We discuss the Stable-Sparse RRT (SST) algorithm \cite{SST}, as well as adaptations and extensions required for our problem setting.

\subsection{Stable-Sparse RRT}

The general framework of SST follows the same principles of other kinodynamic sampling-based algorithms~\cite{LaVKuf01,HauserZ16,FuSSA23}. The SST algorithm was chosen for several reasons. First, an implementation of the algorithm is provided in the state-of-the-art OMPL library~\cite {ompl}. Second, unlike some approaches (e.g.,~\cite{FuSSA23}), the implementation can be applied to various dynamical models and objective functions, including those given by Eqs.~\eqref{dynamics},~\eqref{cost_function}. Third, SST is anytime and asymptotically optimal, i.e., its solution cost keeps improving as additional computational budget is given, and cost of the solution converges to the optimum~\cite{SST}. 

The SST algorithm grows a $\mathcal{T}$ tree embedded in the state space $\mathcal{X}$ of the dynamical model. For the dynamics of the parafoil, $\mathcal{X}\subset\mathbb{R}^4$. In each iteration, the algorithm attempts to extend a tree node. Specifically, SST generates a random state $x_{\text{rand}}\in \mathcal{X}$, and selects the $\mathcal{T}$-tree node $x_{\textup{near}}$ that is closest to $x_{\text{rand}}$. A random control value $\phi\in [-\pi/6,\pi/6]$ and duration $t\in [0,T]$ are then chosen to generate a dynamically feasible trajectory from this node by propagating the dynamics given in Eq.~\eqref{dynamics} using the chosen control value and duration. Given that the trajectory is collision-free, its endpoint is then added to the tree as a new node. The algorithm continues running iterations and searching until a solution is found, or the maximum allowed solution time is reached. The solution with the lowest total cost is then returned.

\subsection{Special Considerations for Human-Piloted Parafoils}

The SST algorithm in our setting optimizes the control effort, which corresponds to the bank angle. The exact expression for the cost \(J\) of node \(n\) is given by:
\begin{equation}
    J_n=J_{n-1}+\phi_n^2 \cdot (t_n-t_{n-1}),
\end{equation}
where \(J_{n-1}\) is the cost of the parent node and \(\phi_n\) is the randomly selected control sampled to generate the trajectory to node \(n\). The cost is chosen to maximize safety for a human pilot, preventing large increases in vertical velocity while ensuring arrival to the landing area at an appropriate height for a landing maneuver. Another change was made to the nearest neighbor search, forcing it to be dependent only on the 3D physical space \((x,y,h)\). Standard SST implementation uses all of the state vector for the nearest neighbor search, but as our workspace is in \(\mathbb{R}^3\), the heading must be excluded from the search.

A key consideration we incorporated into the algorithm is that the parafoil must fly against the wind with minimal turns during landing. Any turns performed during the final approach (last ~\(100[m]\) of height) will lead to a sudden increase in the vertical velocity, which may lead to uncontrolled landings and therefore injury. Such a requirement greatly increases the number of iterations for finding a solution by SST. In particular, in the standard form of the algorithm, it lacks such specialized state-dependent constraints on the control inputs, and instead samples uniform random controls. To circumvent this problem, we use a simple proportional control on the final approach based on the heading and the wind direction. This law calculates the necessary bank angle according to the equation
\begin{equation}
    \phi = 0.5(\psi-\lambda_w), 
\end{equation}
where \(\lambda_w\) is the heading of the wind. This control is used in the propagation instead of the randomly chosen control starting from a chosen threshold height (\(160[m]\) in our case).

\section{Performance Analysis}
\label{results}
In this section, we analyze the performance of the SST algorithm in the context of our problem setup. First, a sample run using a numerical simulation based on the dynamics given in Eq. \ref{dynamics} is presented. In addition, the algorithm's limitations are discussed. Next, the solution quality is analyzed as a function of running time.

The algorithm is implemented using the Open Motion Planning Library (OMPL) \cite{ompl}, which is a state-of-the-art C++ library of commonly used motion planning algorithms and primitives. Our implementation, which is based on SST, was written using the OMPL Python bindings. The OMPL algorithm does not output a solution path directly; rather, it outputs a series of nodes and a control command attached to each node. Results presented in the following subsection are generated by using this control command to propagate the parafoil dynamics using the DOPRI method of the diffrax Python library \cite{diffrax} ODE solver. All simulations were run on a 64-bit laptop with an Intel Core Ultra 7 258V (4.8 GHz) and 32 GB RAM, running Ubuntu 24. The simulations did not rely on GPU acceleration.

\subsection{Sample Run}

\begin{figure*}
\begin{subfigure}{0.5\columnwidth}
\centering
\includegraphics[width=\columnwidth]{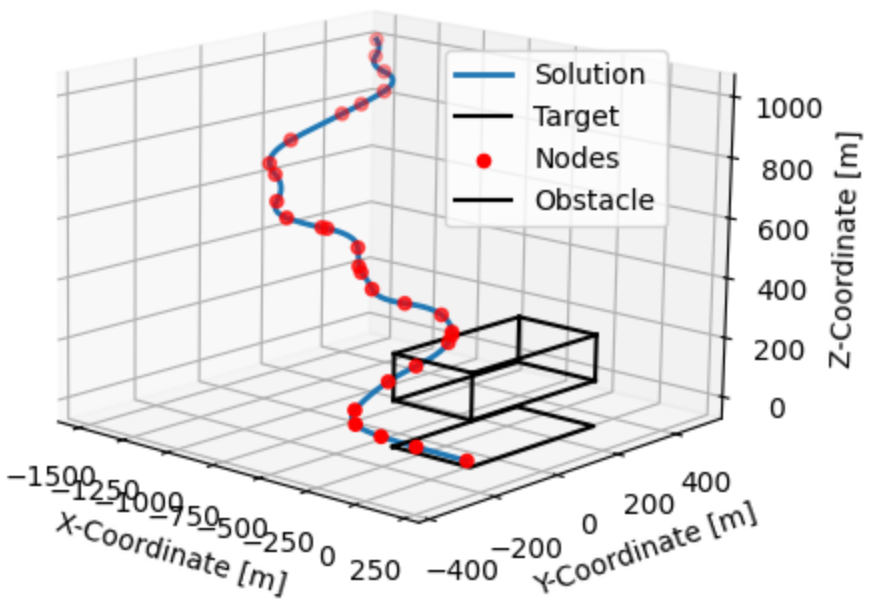}
\subcaption{Trajectory}
\label{exampleCaseTrajec}
\end{subfigure}
\begin{subfigure}{0.5\columnwidth}
\centering
\includegraphics[width=\columnwidth]{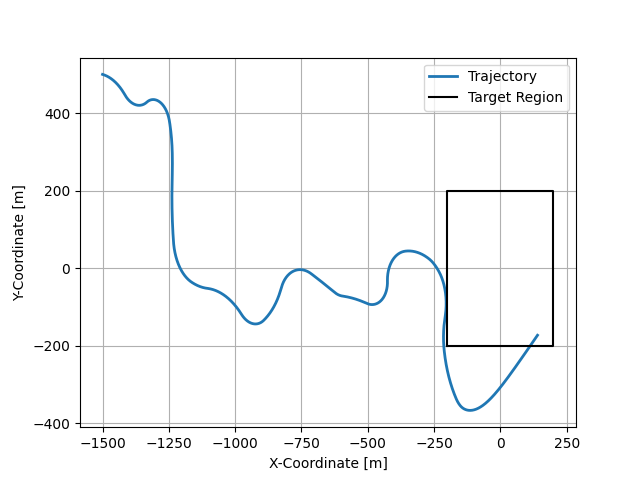}
\subcaption{Top-Down View}
\label{exampleCaseTopDown}
\end{subfigure}
%\newline
\begin{subfigure}{0.5\columnwidth}
\centering
\includegraphics[width=\columnwidth]{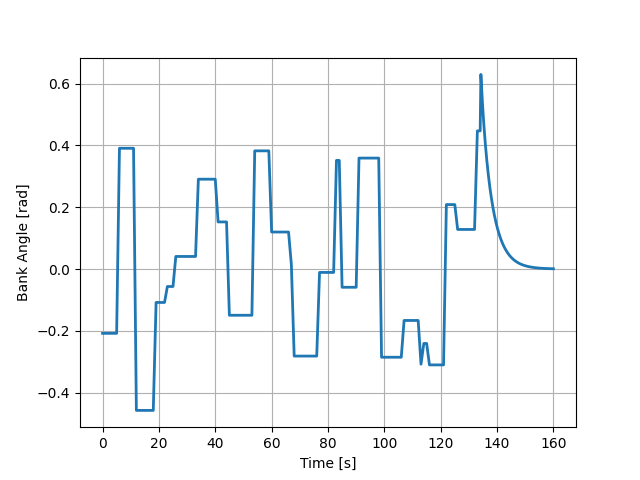}
\subcaption{Bank Angle (Control)}
\label{exampleCaseControl}
\end{subfigure}
\begin{subfigure}{0.5\columnwidth}
\centering
\includegraphics[width=\columnwidth]{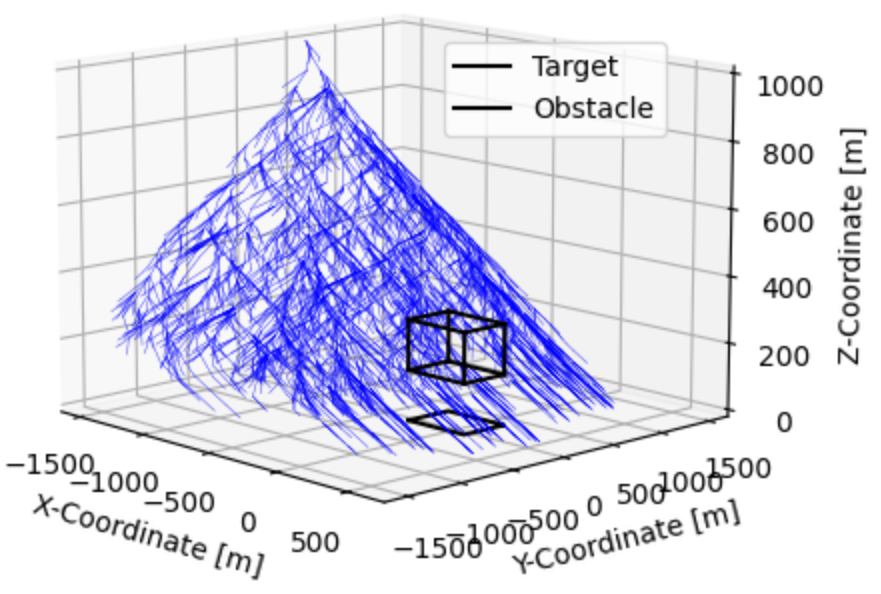}
\subcaption{Tree}
\label{exampleCaseTree}
\end{subfigure}
\caption{Results for a sample run of the algorithm.}
\label{examplecase}
\end{figure*}

To demonstrate the algorithm, a case study is presented in Fig.~\ref{examplecase}. The parafoil starts at the initial conditions \((-1500[m], 500[m], 1000[m], 0[rad])\) with a speed of \(19.2[m/s]\) and a glide ratio of \(3\). The goal region represents a landing site of a \(400[m]\) square with its center at the origin and a height of \(5[m]\). In addition, an obstacle is placed right above the landing area, at a height of \([150[m], 300[m]]\) (human pilots are not supposed to fly above a runway at a certain height). The bounds of the problem are given by a square of size \(3000[m]\) with its center at the origin, and the control (bank angle) is bounded by the value of \(\pi/6[rad]\). The wind vector is set to be \((-3[m/s],-3[m/s],0[m/s])\). The duration step size is \(1[s]\) with a maximum number of randomly sampled steps for each propagation of \(10\), and the solution running time is set to  \(40[s]\).

The algorithm successfully finds a solution trajectory (Fig. \ref{exampleCaseTrajec}) that obeys the constraints of the problem, namely, obstacle avoidance and landing against the direction of the wind (Fig. \ref{exampleCaseTopDown}). For most of the flight, the control is discrete, representing the randomly sampled control value at each node. The discrete control may pose a challenge for real-world implementation and may require further research to achieve the desired solution. The controlled final approach can be seen towards the end of the flight (Fig. \ref{exampleCaseControl}), where the control switches from discrete values to a smooth, continuous function that decreases in time. This is the effect of the proportional gain discussed earlier, and shown in Fig. \ref{exampleCaseTree} as well: the bottom part of the exploration tree points against the wind.

\subsection{Sensitivity Analysis}

Next, we performed a sensitivity analysis on various problem parameters with a maximum solution time of \(30[s]\), a parafoil speed of \(15[m/s]\), and a glide ratio of $3$. Flight-day uncertainties were chosen for the analysis; these are the parameters that remain unknown until the flight starts (i.e. initial location, initial heading, and wind parameters). For each set of parameters, the algorithm was invoked 200 times.

First, we ran the analysis on both the initial location and the initial heading, and neither altered the success rates. With respect to the wind direction, we observed minimal dependence---for large differences between the direction to the target and the direction of the wind, the chances of success were slightly lower. For wind magnitude, a larger impact was seen, with the algorithm having a good chance of successfully finding a solution for wind magnitudes of \(8[m/s]\) and under. Going above \(8[m/s]\) greatly decreases the chances of finding a solution. This may be due to the wind overpowering the velocity of the parafoil and physically preventing the parafoil from arriving at the target location. This implies that the cutoff presented here is relevant only to the specific conditions used in the analysis and may change with parafoil speed. Certainly, as will be seen in the next section, increasing the maximum solution time will always increase the chances of finding a solution.

\subsection{Runtime Analysis}
\begin{figure}[h]
\centering
\includegraphics[width=0.99\linewidth]{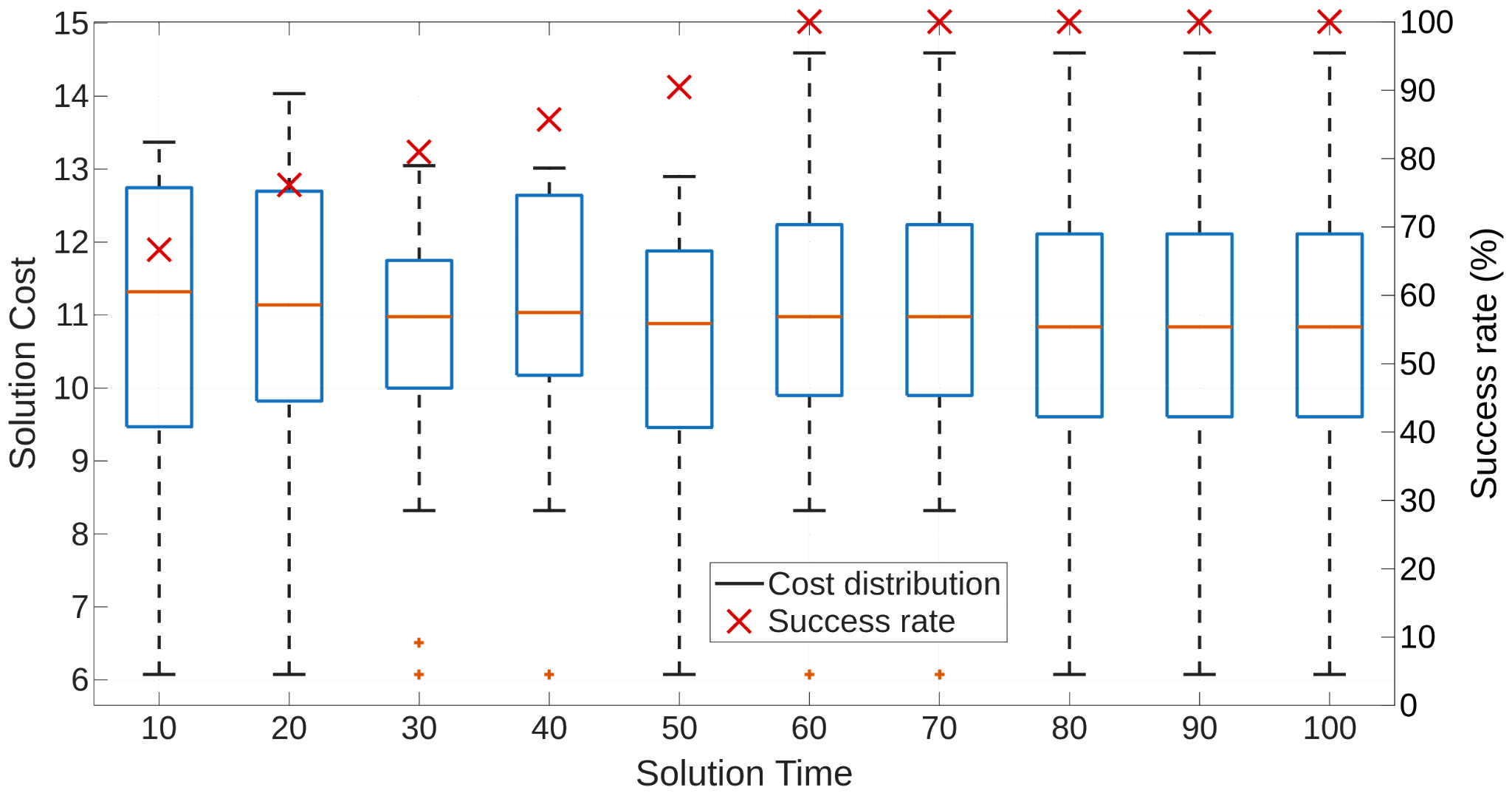}
\caption{Final solution cost for different runtimes. The plot follows standard notation for a box plot.}
\label{runtimeanalysis}
\end{figure}

Next, a runtime analysis is performed for solution times up to \(100 [s]\) with intervals of \(10 [s]\) and the same parafoil conditions as in the last section. Twenty runs of the algorithm were invoked, and then the cost of the currently found solution was sampled every ten seconds. The results are shown in Fig. \ref{runtimeanalysis}.

For runtime above \(50[s]\) the algorithm consistently returns a solution, but there is a plateau in the solution cost vs. runtime. This plateau is most likely due to the algorithm finding the best solution around this time. It can also be seen that the success rate reaches $100\%$ at this time as well. We note that  some runs of the algorithm took a fairly long time to find a solution. This may pose a barrier for future work (such as multi-agent solutions) and may need to be addressed in future work. 

\section{Real Flight Comparison}
\label{realdata}

To test the algorithm's stated goal of providing a solution for human-piloted flight, the algorithm's solution was compared to real-world data. After discussing below the data collection methodology, we run several comparisons between the data and the algorithm's solution.

For the sake of discussion, we define here a new attribute: the relative optimality gap (ROG), which measures the gap between the human performance and the optimal performance (which is approximated by the solution of the algorithm). Specifically, for a given scenario, denote by $c_h(t)$ and $c_a(t)$ the cumulative cost (Eq.~\eqref{cost_function}) for the human data and the algorithm solution, respectively, up to (normalized) time $t\in [0,1]$. Then, the ROG at time \(t\) is given by $\frac{c_h(t)-c_a(t)}{c_h(1)}$. That is, positive values indicate that the algorithm obtained a better solution up to a given point $t$.

Overall, we show that the algorithm performs better than a human pilot while minimizing the control effort. Although a human pilot tends to fly straight to the general vicinity of the landing area, the algorithm takes a smoother, more gradual descent, requiring less control effort, leading to improvements of up to $0.85$ in ROG.

\subsection{Data Collection}

The flight data was gathered using a Flysight 1 GPS sensor attached to a \(135[ft^2]\) RAM-air parafoil manufactured by Innovative Parachute Technologies (IPT). The parachute is a semi-elliptical, zero-porosity, 9-celled model, called Axon (see Fig. \ref{annapicture}), and is loaded\footnote{The wing loading is the ratio between the skydiver's weight [$lb$] and the wing surface area [$ft^2$].}  around \( [1.0,1.1] \). A total of 19 flights were performed, with a typical duration of \(150-200[s]\) and starting altitudes of \(900-1000[m]\). The GPS sensor returns the time of each measurement, the location of the parafoil (latitude, longitude, altitude), the velocity vector (north, east, down), the heading relative to North, the accuracy of each measurement, the GPS fix type, and the number of satellites detected by the sensor. After GPS data is collected, it is transferred to the NED (North-East-Down) coordinate system with the landing point as the origin.

\subsection{Example Comparison}

\begin{figure*}[h!]
\begin{subfigure}{\columnwidth}
\centering
\includegraphics[width=\columnwidth]{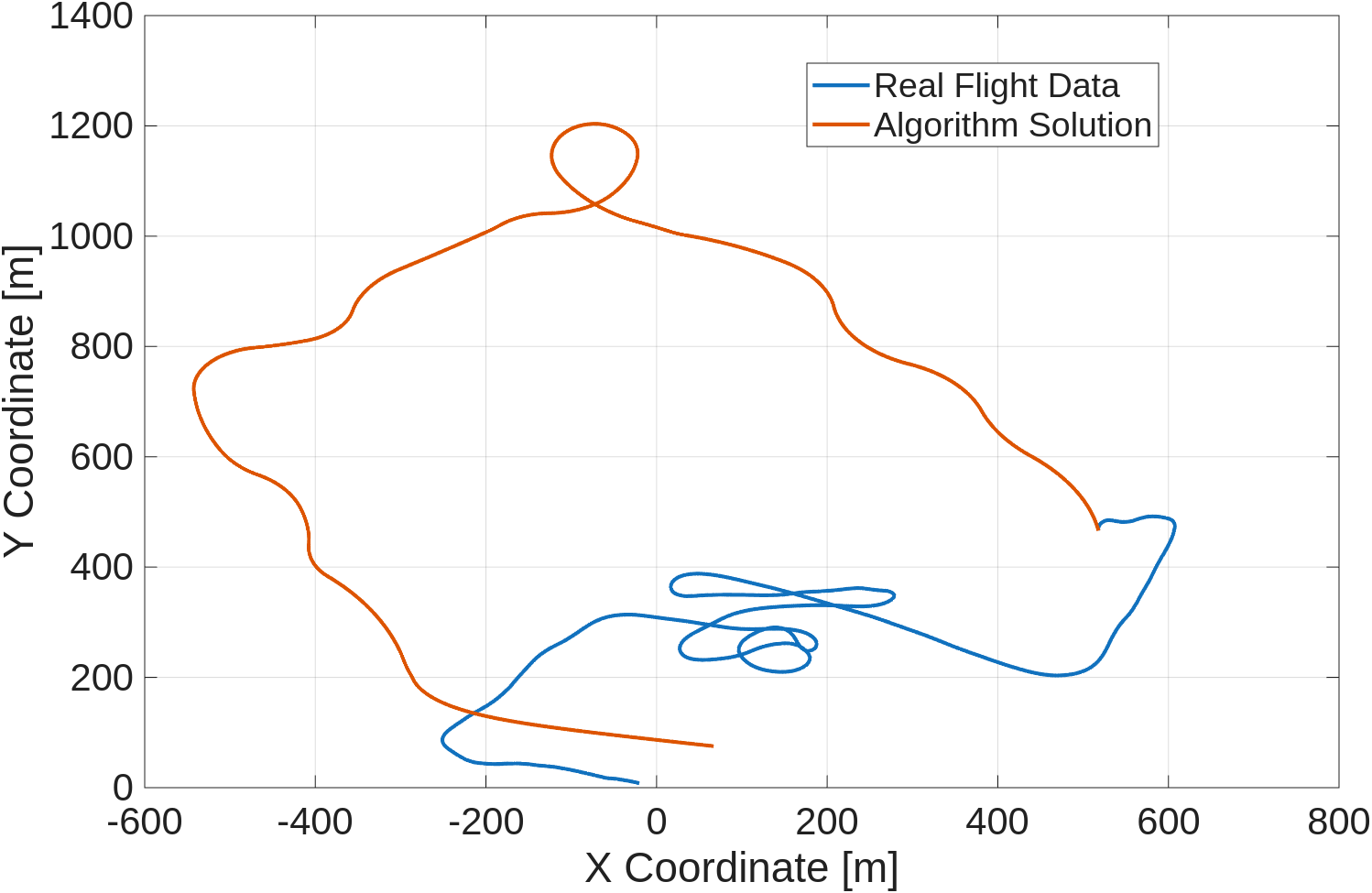}
\subcaption{Top down View}
\label{realworld2d}
\end{subfigure}
\begin{subfigure}{\columnwidth}
\centering
\includegraphics[width=\columnwidth]{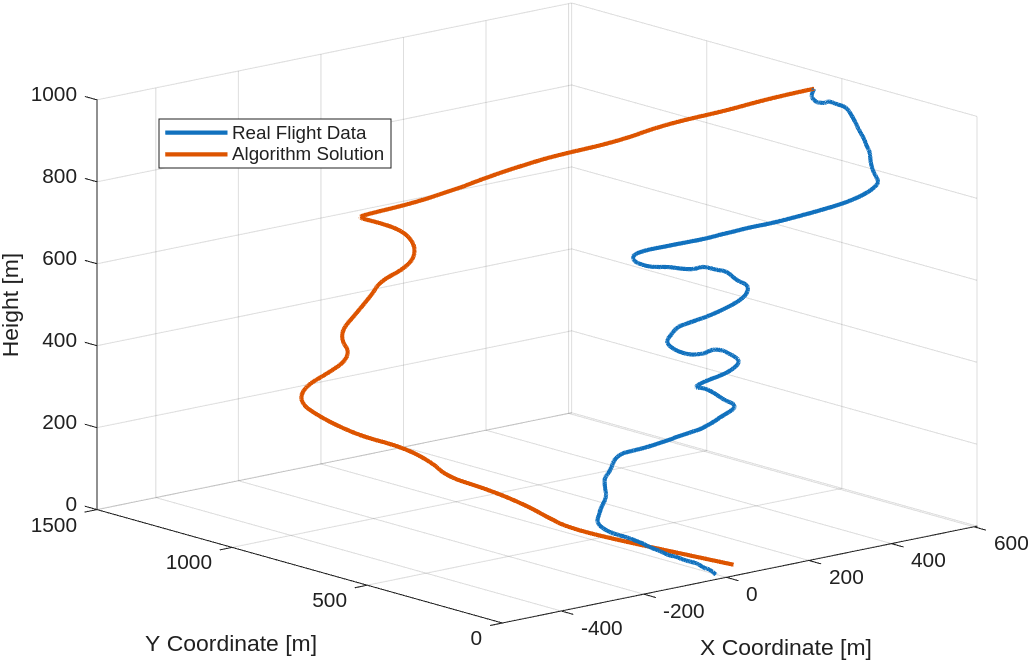}
\subcaption{3D View}
\label{realworld3d}
\end{subfigure}
\caption{Example case of real-life data compared to the algorithm solution.}
\label{realworldcomp}
\end{figure*}

A comparison of the algorithm's solution to a real flight case is shown in Fig. \ref{realworldcomp}. The algorithm receives initial conditions and wind conditions (an estimated average wind vector assumed to be constant for the whole flight) provided by the real data, as well as a glide ratio of \(3\) and a speed of \(15[m/s]\), and proceeds to generate a solution.

Note that both the algorithm and the human pilot land against the wind in the same direction (Fig. \ref{realworld2d}). However, the maneuver into the final approach occurs at different heights, and therefore, the algorithm's solution continues for a longer time against the wind than the flight data (Fig. \ref{realworld3d}), avoiding the danger of a low-height turn.

The different approaches between the algorithm and the human pilot can be seen here quite clearly. The algorithm prefers a smooth and gradual descent with minimal turns, minimizing the control effort as defined in the cost function. On the other hand, a human pilot generally prefers to fly to the vicinity of the landing area, reduce altitude with  a spiral maneuver, and  enter the landing pattern. While this behavior may be optimal in the short term, it cannot lead to a globally optimal solution due to the additional costs incurred with the spiral maneuver. This behavior leads to a significant advantage for the algorithm solution, with a ROG of $0.58$.

\subsection{Cost Comparison}

\begin{figure}[h!]
\centering
\includegraphics[width=0.99\linewidth]{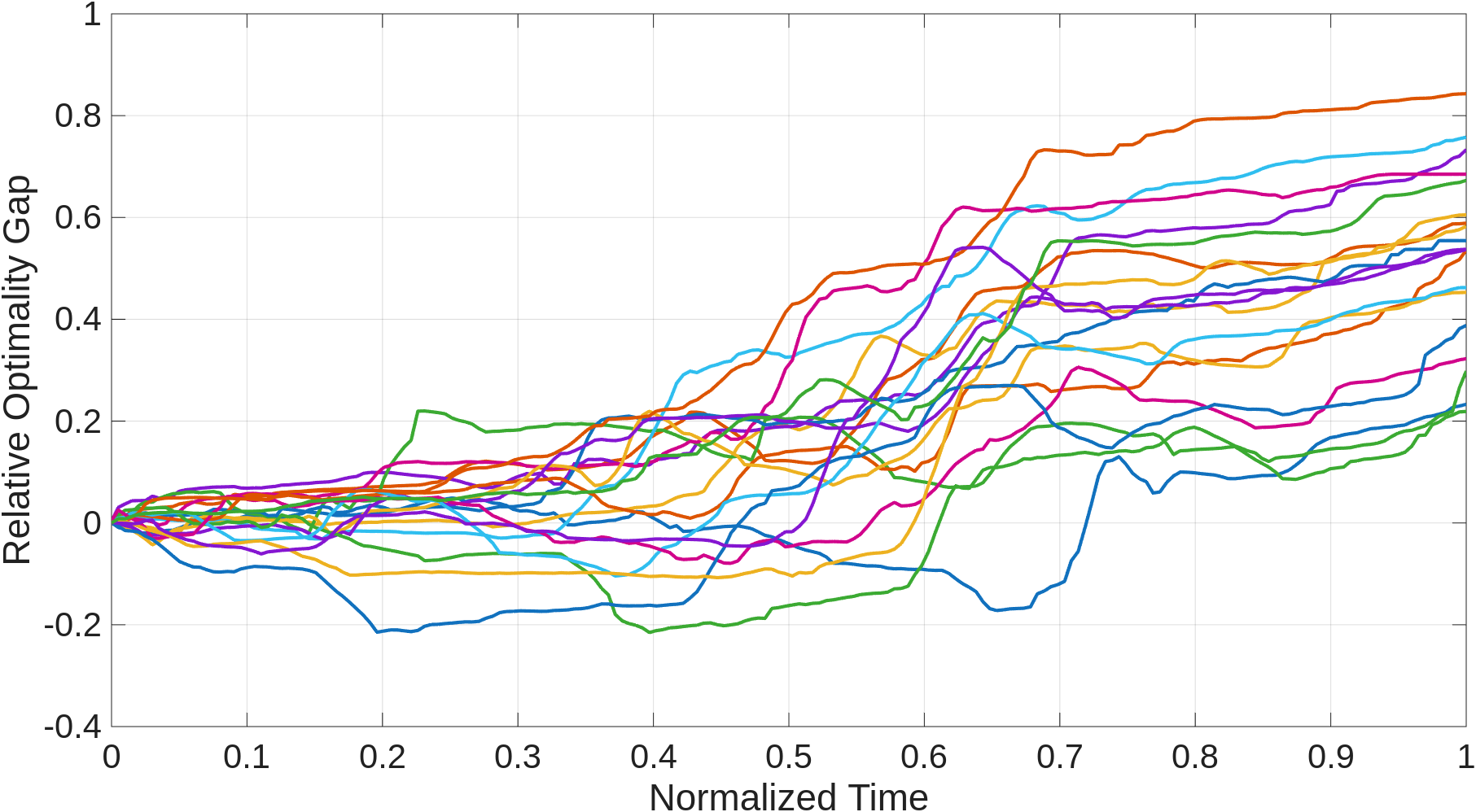}
\caption{ROG between the algorithm solution and the flight data for 19 different datasets (a distinct color per data set).}
\label{realDataCost}
\end{figure}

To obtain a quantitative analysis of the cost, ten runs were performed for each of the 19 scenarios of real flight data. The run with the lowest cost (best algorithm performance) was then selected from the ten. The ROG vs. normalized time across all 19 datasets is shown in Fig. \ref{realDataCost}.

First, observe that most of the time, the algorithm performs better than a human when minimizing the cost. Moreover, in all cases displayed in Fig. \ref{realDataCost}, at the completion time, the algorithm obtains a ROG in the range $[0.2,0.85]$, which highlights the substantial optimality gap of the human pilot. A general trend in the ROG characterized by an initial levelness, then a sharp increase, and a final return to levelness can be seen in all of the datasets. This trend can be explained though the different approaches taken by the human pilot and the algorithm. In the beginning the human pilot generally flies straight towards the landing area, producing a negative ROG (i.e. a better performance), while the algorithm starts turning immediately. However, globally the human's approach is suboptimal since the spiral maneuver later on in the flight results in  the sharp increase of the cost. Finally, in the last section of the flight both human and virtual pilots fly against the wind to perform landing, hence producing a similar cost. In this way, the smooth and gradual descent of the algorithm is the globally optimal solution, producing the final ROG values seen in Fig. \ref{realDataCost}.

\subsection{Lowest Algorithm Performance Case}
\begin{figure}[h]
\centering
\includegraphics[width=0.99\linewidth]{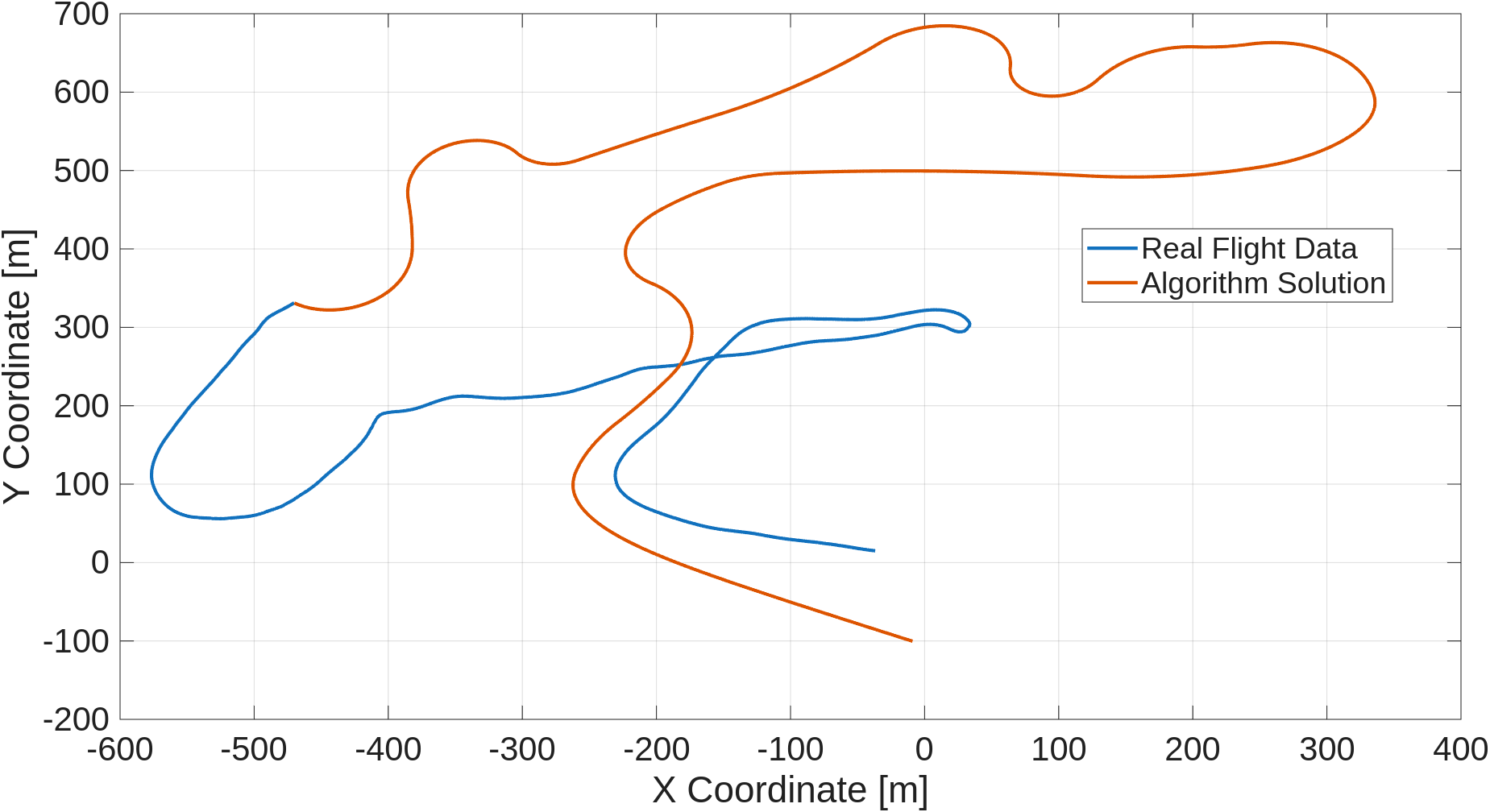}
\caption{Top-down view of the lowest cost case trajectory.}
\label{lowestcase2D}
\end{figure}

In this section, the case that results in the lowest difference between the algorithm performance and the human pilot at the end of the flight is analyzed to gain further insight into the differences between human and algorithmic behavior (see Fig. \ref{lowestcase2D}). The algorithm outperforms the human pilot with an ROG of $0.16$ for the solution shown in Fig. \ref{lowestcase2D}.

As noted earlier, the algorithm generates a trajectory with a larger horizontal span to minimize cost. The human pilot, however, makes an initial U-turn before heading straight to the horseshoe-like landing maneuver, which is also entered with a U-turn. The human pilot decides to fly straight against the wind as this is the direction towards the landing area, while the algorithm finds a better solution in the global sense. This straight flight against the wind is most likely done to ensure that the human pilot arrives at the landing area with enough height for the landing maneuver, since opening of the parafoil downwind from the landing area may create apprehension. The algorithm has no such pitfalls, and finds a longer trajectory with a more gradual descent even though it starts from the same initial position. Smoothing the generated trajectory may further improve solution cost and will be explored in future work.

\subsection{Specific Constraint Comparison}
\label{elsinoresec}

\begin{figure}[b]
\centering
\includegraphics[width=0.8\linewidth]{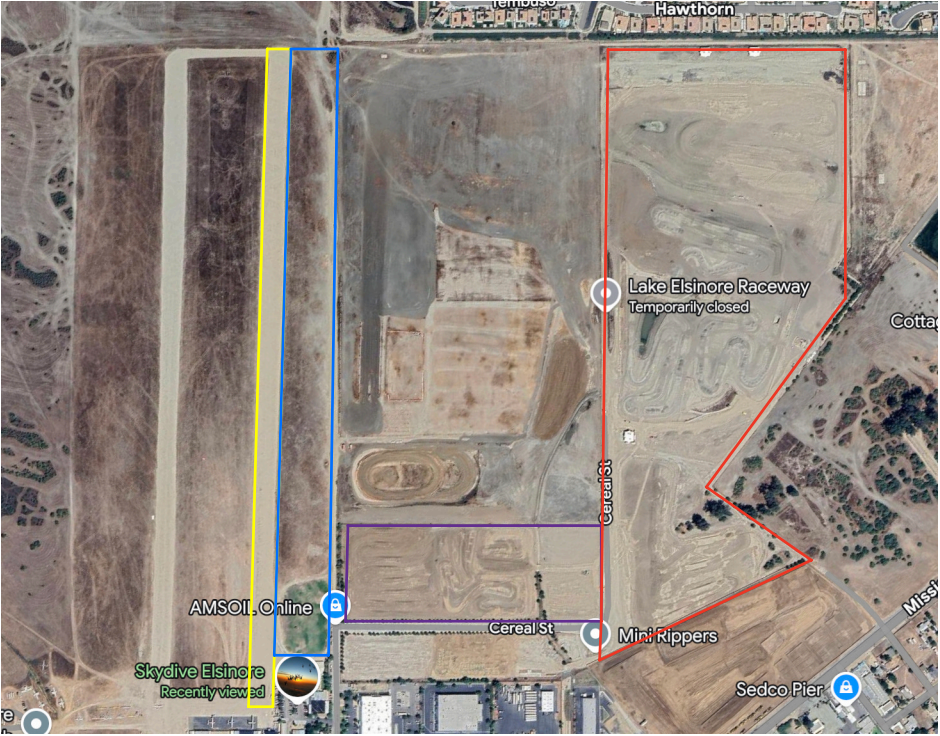}
\caption{Map of the Elsinore Skydive site.}
\label{elsinoremap}
\end{figure}

In the previous experiments, we did not explicitly account for terrain-related constraints (see the additional discussion on limitations below). In this experiment, we integrate those constraints to assess their impact on the algorithm solution, and to provide a more accurate comparison. Specifically, we consider a data set for the Elsinore Skydive site in Lake Elsinore, California, which was modeled and presented to the algorithm for solution (Fig. \ref{elsinoremap}). The blue area is the landing area of the parafoil, while the yellow area is the runway. Flying above the runway below \(1500 [ft]\) (\(457 [m]\)) is forbidden due to the airplane traffic. The red marked area is the Elsinore Raceway,  flying above it is undesirable since this area is totally inappropriate for landing (in case the parafoil experiences a malfunction and is forced to lose altitude faster than expected). The purple-marked area is another raceway, with flying being permitted here only on the way to landing, under \(1000[ft]\) (\(300[m])\). 

\begin{figure}[h!]
\centering
\includegraphics[width=0.8\linewidth]{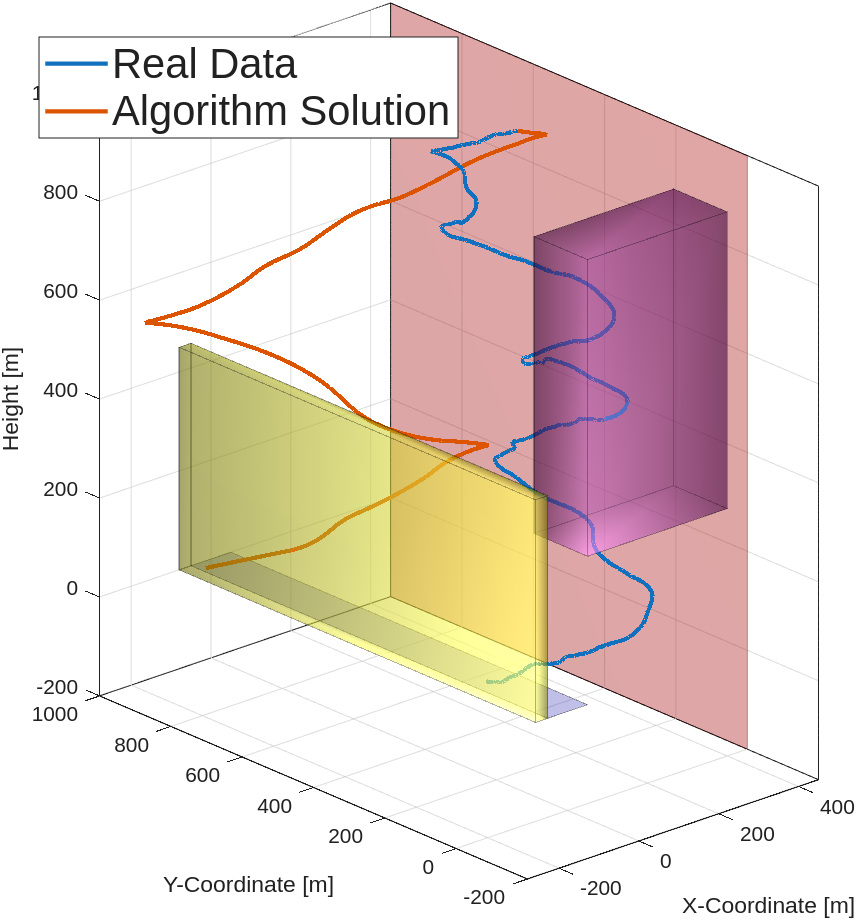}
\caption{3D view for the Elsinore Skydive site trajectory.}
\label{elsinore3D}
\end{figure}
\begin{figure}[h!]
\centering
\includegraphics[width=0.99\linewidth]{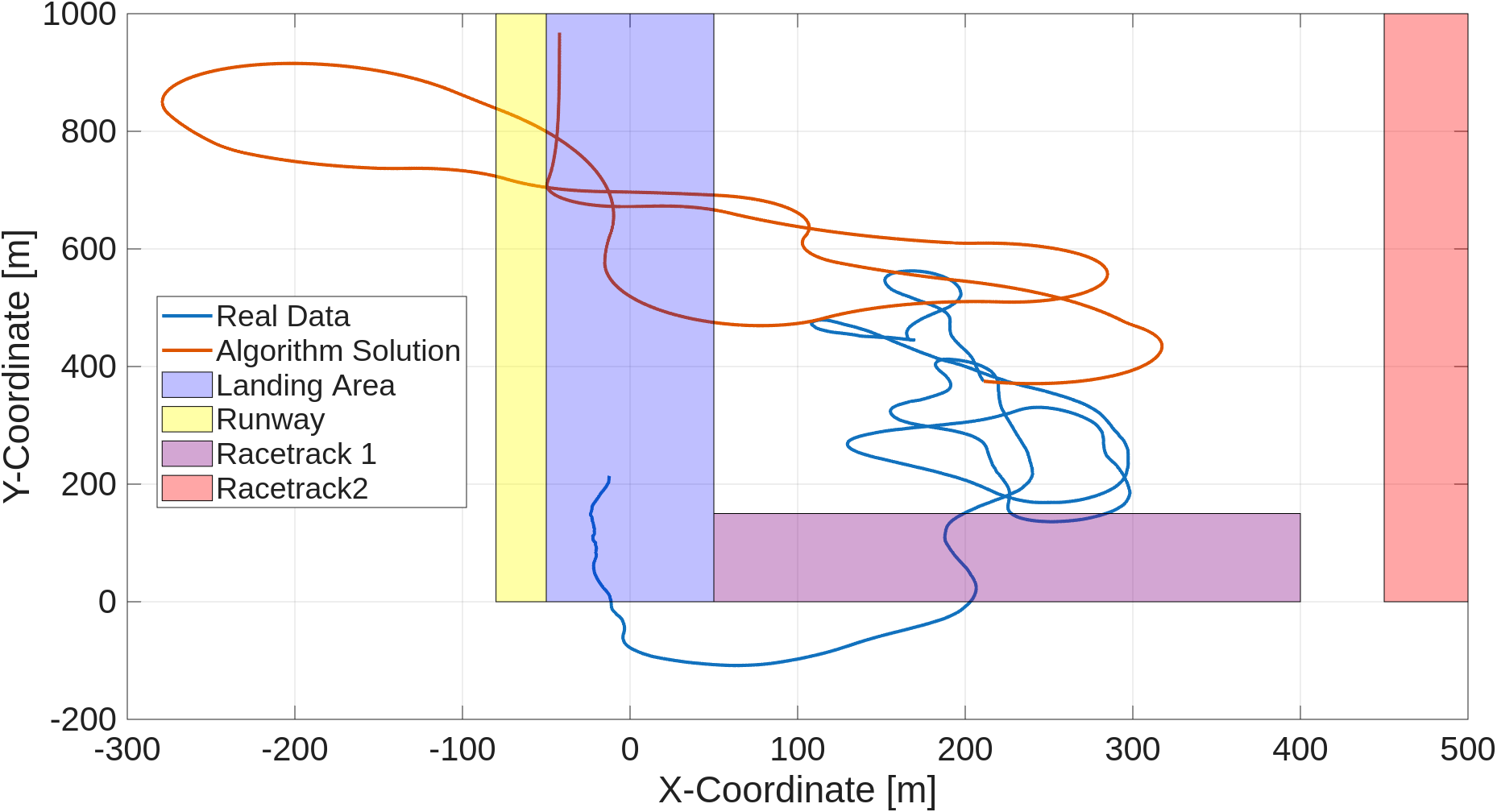}
\caption{Top-down view of the Elsinore Skydive site trajectory.}
\label{elsinore2D}
\end{figure}

It can be seen that even in a tight environment where the human pilot is forced to spiral, the algorithm finds a smoother way to land. The first big difference, evident in Figs. \ref{elsinore3D}-\ref{elsinore2D}, is that the algorithm flies on both sides of the runway and crosses it twice, while staying above the maximum height limit. The human pilot 
stays on one side of the runway at all times, since estimating the altitude required to fly across and return without violating the constraint is a difficult task. However, spiral maneuvers can still be seen in the algorithm's descent in Fig. \ref{elsinore2D}, although they are much wider and smoother than the small spirals of the human pilot. This may be because the human pilot stays farther away from the obstacles due to her apprehension of violating the constraints. Finally, the human pilot lands much closer to the Y-axis origin, even though this involves dealing with the purple obstacle. This is due to the  proximity of the skydiving center, located close to the Y-axis origin: the human pilot would prefer to walk back less after landing. Even with all the above constraints, the algorithm still performs better than the human pilot, achieving the ROG of  $0.7$.

\subsection{Key Insights}

The above analysis reveals a strategical difference in trajectory planning between the human pilot and the algorithm: the algorithm simultaneously and proportionally utilizes all the three dimensions of the available space, whereas humans are not accustomed to plan in 3D and tend to plan the flight in the horizontal plane first and make vertical adjustments later, upon arrival at the desired horizontal position. Such a decoupling approach cannot be optimal, and can result in running out of altitude during the first stage, or performing a spiral maneuver during the second stage.  Spiral maneuvers may be dangerous due to the increase in vertical velocity, and are forbidden in some locations. For these reasons the algorithm provided lower cost and safer trajectories in all examined scenarios. Therefore, our future work will focus on utilizing the algorithm to aid novice parafoil pilots plan their path in 3D, so that they can develop a qualitatively different flying strategy.

\subsection{Limitations}

The algorithm has a number of limitations that are planned to be addressed in future work. First, no other parafoils or obstacles are present in the data except for the specific modeling done for Sec. \ref{elsinoresec}. Presence of other parafoils or obstacles may affect human pilot decision making and require collision avoidance maneuvers. However, the data presented above was collected during flights with very few  obstacles and other parafoils, in order to ensure that their impact is minimal. Another limitation is the structure of the wind model. Winds modeled here are constant vectors, and in future work more complicated variable wind models are planned to be added. In the same vein, the parafoil dynamics chosen are quite simple and more complicated dynamics may be added in the future, possibly taking into account turbulence from other parafoils in the air. New dynamics may also include a control model for the pilot inputs instead of a straight bank angle control.

\section{Conclusions}
\label{conc}

In this work, we studied the performance of a human-piloted parafoil coming in for a landing, as compared to an asymptotically-optimal sampling-based algorithm. While human pilots prefer to remain close to the landing area and reduce altitude through spiral maneuvers, the algorithm prefers a smoother and gradual descent, resulting in a lower cost. Such an algorithm can be used in conjunction with a flight simulator to train parafoil pilots prior to jumping, providing novice pilots with valuable experience and potentially reducing injuries.

Additional research is required with regards to the presence of the spiral maneuver in the human piloted flight and its absence in the algorithm solution. As stated above, this maneuver may be due to factors not included in the algorithm solution, such as wind estimation, avoidance of other parafoils, validation of parafoil functionality. Further research shall seek to implement these variables into the algorithm to investigate their effects on the solution and determine conditions for the emergence of spiral maneuvers. Future work will also focus on expanding the problem to include multiple agents converging to the same landing area, as well as introducing more accurate wind and parafoil models. An equally important aspect of future work is conveying the paper's discoveries to a general skydiving audience.  

\bibliographystyle{unsrt}
\bibliography{references}
\end{document}